\documentclass[10pt,twocolumn,letterpaper]{article}

\usepackage{wacv}              

\usepackage{multirow, makecell}
\usepackage{multirow}
\usepackage{booktabs}

\usepackage{graphicx}
\usepackage{algorithm}
\usepackage{algorithmic}
\usepackage[utf8]{inputenc} 
\usepackage[T5]{fontenc}     
\usepackage{caption} 

\usepackage[table]{xcolor}

\definecolor{wacvblue}{rgb}{0.21,0.49,0.74}
\usepackage[pagebackref,breaklinks,colorlinks,allcolors=wacvblue]{hyperref}

\def\wacvPaperID{3311} 
\def\confName{WACV}
\def\confYear{2026}

\title{PADM: A Physics-aware Diffusion Model for Attenuation Correction}

\author{
Trung Kien Pham$^{1}$\thanks{Co-first authors (Equal contribution).}\hspace{5pt}, Hoang Minh Vu$^{1*}$, Anh Duc Chu$^{1*}$, Dac Thai Nguyen$^{1*}$, Trung Thanh Nguyen$^{2}$, \\ Thao Nguyen Truong$^{3}$, Mai Hong Son$^{4}$, Thanh Trung Nguyen$^{4}$, and Phi Le Nguyen$^{1}$\thanks{Corresponding author: \texttt{lenp@soict.hust.edu.vn}.}\\
$^{1}$AI4LIFE, Hanoi University of Science and Technology, Vietnam;
$^{2}$Nagoya Univeristy, Japan \\ 
$^{3}$National Institute of Advanced Industrial Science and Technology, Japan\\
$^{4}$108 Military Central Hospital, Vietnam
}

\begin{document}

\maketitle
\begin{abstract}
Attenuation artifacts remain a significant challenge in cardiac Myocardial Perfusion Imaging (MPI) using Single-Photon Emission Computed Tomography (SPECT), often compromising diagnostic accuracy and reducing clinical interpretability. While hybrid SPECT/CT systems mitigate these artifacts through CT-derived attenuation maps, their high cost, limited accessibility, and added radiation exposure hinder widespread clinical adoption. In this study, we propose a novel CT-free solution to attenuation correction in cardiac SPECT. Specifically, we introduce \textbf{P}hysics-aware \textbf{A}ttenuation Correction \textbf{D}iffusion \textbf{M}odel (PADM), a diffusion-based generative method that incorporates explicit physics priors via a teacher--student distillation mechanism. This approach enables attenuation artifact correction using only Non-Attenuation-Corrected (NAC) input, while still benefiting from physics-informed supervision during training. To support this work, we also introduce CardiAC, a comprehensive dataset comprising 424 patient studies with paired NAC and Attenuation-Corrected (AC) reconstructions, alongside high-resolution CT-based attenuation maps. Extensive experiments demonstrate that PADM outperforms state-of-the-art generative models, delivering superior reconstruction fidelity across both quantitative metrics and visual assessment. 
\end{abstract}

\section{Introduction}

Myocardial Perfusion Imaging (MPI) using Single Photon Emission Computed Tomography (SPECT) is a widely employed, non-invasive imaging modality for the diagnosis, risk stratification, prognostication, and therapeutic management of patients with coronary artery disease. 
By acquiring perfusion slices along three orthogonal anatomical planes, i.e., short axis, vertical long axis, and horizontal long axis, MPI provides clinicians with essential insights into myocardial blood flow, enabling the evaluation of cardiac function, perfusion abnormalities, and tissue viability.

Despite its clinical utility, the diagnostic performance of SPECT - MPI is often compromised by attenuation artifacts, spurious image distortions arising from heterogeneous tissue densities. 
These artifacts can obscure true perfusion defects or simulate false positives, leading to reduced specificity and diagnostic confidence. The impact of attenuation is particularly pronounced in obese patients and those with subdiaphragmatic anatomical structures such as bowel loops or a raised diaphragm, which significantly affect photon transmission paths~\cite{Grossman2004}.

To mitigate these limitations, several artifact-reduction strategies have been adopted in clinical practice. Techniques such as ECG-gated MPI and prone positioning aim to reduce motion-related and positional attenuation; however, their efficacy remains inconsistent and heavily patient-dependent~\cite{Hayes1633}. A more robust solution lies in hybrid SPECT/CT systems, which incorporate low-dose computed tomography to estimate voxel-wise attenuation coefficients. 
These attenuation maps are integrated into the reconstruction algorithm to generate Attenuation-Corrected (AC) images with improved diagnostic fidelity~\cite{Huang2011, Goetze1090, Saleki2019}. Nevertheless, widespread adoption of this approach is constrained by high system costs, increased radiation exposure, and limited accessibility, especially in resource-limited clinical environments.

To address these limitations, several methods have emerged as an alternative route for generating Attenuation-Corrected (AC) images directly from Non-Attenuation-Corrected (NAC) inputs. 
Existing approaches for this problem can be broadly categorized into two groups: neural network-based~\cite{goodfellow2014generative, isola2017image, zhu2017unpaired, li2023bbdm} and formula-based~\cite{Huang2011, Goetze1090, Saleki2019}. The former formulates attenuation correction as an image-to-image translation task. Deep generative models, such as Generative Adversarial Networks (GANs)~\cite{goodfellow2014generative} or diffusion models~\cite{croitoru2023diffusion}, are trained to map NAC inputs to synthetic AC outputs. While these methods can learn the underlying distribution of AC images and produce visually compelling results, they often lack physical interpretability. Consequently, generated images may deviate from the ground-truth physics, undermining their clinical reliability. Formula-based approaches~\cite{Huang2011, Goetze1090, Saleki2019}, by contrast, use explicit physical modeling by estimating attenuation maps from CT scans and applying them to correct NAC images through standard reconstruction pipelines. While grounded in well-established imaging physics, these methods are often limited by their inflexibility and inability to generalize across diverse anatomical and pathological variations. Furthermore, they inherently require access to CT input during both training and inference, perpetuating the same cost and accessibility issues they aim to resolve.

To bridge the gap between these two paradigms, we propose a hybrid framework that combines the data-driven strengths of deep generative models with the rigor and interpretability of physics-based modeling. Specifically, we introduce PADM, a novel Physics-aware Attenuation Correction Diffusion Model for cardiac SPECT attenuation correction.
PADM introduces three core innovations to enable accurate attenuation correction without requiring CT input at inference. First, it employs a diffusion-based generative model to iteratively refine NAC inputs into high-fidelity AC images. Second, it incorporates physics-guided conditioning using CT-derived attenuation maps during training, allowing the model to learn physically meaningful corrections. Finally, it leverages a knowledge distillation framework, where a CT-informed teacher transfers its expertise to a NAC-only student model, ensuring CT-free deployment without compromising accuracy.
To further advance research in this area, we also contribute CardiAC, a comprehensive dataset comprising 424 patient studies with paired NAC and AC reconstructions, along with high-resolution CT-based attenuation maps. 
To the best of our knowledge, no publicly available dataset currently exists for NAC-to-AC reconstruction in cardiac SPECT. We will release the CardiAC dataset to support research in this area.

\noindent The main contributions of this study are as follows:
\begin{itemize}
    \item We propose PADM, a diffusion-based generative model that integrates Physics-based supervision into the learning process. 
    PADM's teacher--student architecture enables accurate, CT-free attenuation correction at inference.
    \item We introduce CardiAC, a comprehensive dataset for cardiac SPECT attenuation correction. 
    CardiAC offers high-resolution imaging and broad clinical diversity, establishing a strong benchmark for future research.
    \item We conduct comprehensive evaluations against state-of-the-art generative baselines. PADM demonstrates consistent improvements in both quantitative performance metrics and perceptual quality, validating the effectiveness of combining physical priors with advanced diffusion modeling for cardiac SPECT attenuation correction.
\end{itemize}

\begin{table*}[t]
\centering
\caption{\textbf{Comparison of existing cardiac SPECT datasets with paired NAC and AC reconstructions.} HLA = Horizontal Long-Axis, VLA = Vertical Long-Axis, SA = Short-Axis. $\checkmark$ = available; n/r = not reported; ``ext.'' = external test studies.}
\small
\begin{tabular}{lccccccc}
\toprule
\textbf{Dataset} & \textbf{Protocol} & \textbf{Vol. Size (H$\times$W$\times$Slices)} & \textbf{\# Studies} & \textbf{HLA} & \textbf{VLA} & \textbf{SA} \\
\midrule
Shanbhag et al.~\cite{shanbhag2022deepac} & Stress+Rest & n/r & 4,886 (+604 ext.) & n/r & $\checkmark$ & $\checkmark$ \\
Yang et al.~\cite{yang2025ctfreeac}     & Stress+Rest & 64 $\times$ 64 $\times$ 32 & 202 & $\checkmark$ & $\checkmark$ & $\checkmark$ \\
Chen et al.~\cite{chen2022durdn}                 & Stress+Rest & 32 $\times$ 32 $\times$ 32 & 172 & $\checkmark$ & $\checkmark$ & $\checkmark$ \\
Mostafapour et al.~\cite{mostafapour2022mpiac}  & Stress+Rest & 64 $\times$ 64 $\times$ 40 & 99  & $\checkmark$ & $\checkmark$ & $\checkmark$ \\
Arabi \& Zaidi~\cite{abi2022mpsmap} & Stress+Rest & 64 $\times$ 64 $\times$ 32 & 345 & $\checkmark$ & $\checkmark$ & $\checkmark$ \\
Torkaman et al.~\cite{torkaman2022cgan}  & Stress-only  & 64 $\times$ 64 $\times$ 32 & 100 & n/r & n/r & n/r \\
Yang et al.~\cite{yang2022stress70} & Stress-only  & 70 $\times$ 70 $\times$ 50 & 100 & n/r & n/r & n/r \\
\cmidrule(lr){1-7}
\textbf{CardiAC (Ours)} & Stress+Rest & {128 $\times$ 128 $\times$ $D$} ($25 \leq D \leq 49$) & {424} & $\checkmark$ & $\checkmark$ & $\checkmark$ \\
\bottomrule
\end{tabular}
\label{tab:spect_ac_datasets_compact}
\vspace{-3pt}
\end{table*}


\section{Related Work}

\subsection{Paired NAC and AC Datasets}
\label{subsec:dataset}
Recent advances in CT-free attenuation correction for myocardial perfusion SPECT have been supported by the emergence of datasets containing paired NAC and AC reconstructions. Table~\ref{tab:spect_ac_datasets_compact} summarizes representative datasets in this domain. 
The largest to date is the dataset from Shanbhag et al.~\cite{shanbhag2022deepac}, comprising 4,886 studies collected from Yale University, with an additional 604 external cases from University of Zurich and University of Calgary. 
Other datasets~\cite{yang2025ctfreeac, mostafapour2022mpiac, chen2022durdn, torkaman2022cgan, abi2022mpsmap, yang2022stress70} are more limited in scale, containing from 99 to 345 studies, and often suffer from low spatial resolution or incomplete anatomical orientation coverage. 
Some are further restricted to stress-only protocols and lack full-axis orientation support.
In contrast, the proposed CardiAC dataset provides 424 studies with high-resolution 128$\times$128 volumes with complete axis-aligned NAC and AC image pairs under stress and rest conditions, offering high dataset volume, anatomical completeness, and reconstruction quality.


\subsection{Attenuation Correction of SPECT Images}
\label{subsec:attenuation_correction}
Attenuation correction in SPECT is traditionally performed using iterative reconstruction methods, which leverage forward and backward projections, often guided by CT-derived attenuation maps~\cite{Beister2012, Liu2014}. 
While effective, IR-based techniques are computationally intensive, require access to CT hardware, and are prone to artifacts or misalignment~\cite{Goetze1090, Bockisch2009}, motivating CT-free alternatives. 
Recent advances in deep learning have enabled data-driven AC approaches that bypass the need for CT input by learning direct mappings from NAC to AC images. Generative models like MedGAN~\cite{armanious2020medgan} and SynDiff~\cite{ozbey2023unsupervisedmedicalimagetranslation} have shown promising results in modality translation, yet many DL-based methods are limited by architecture simplicity and the scarcity of high-quality paired NAC and AC datasets~\cite{shanbhag2022deepac, mostafapour2022mpiac}.

\subsection{Image-to-Image Translation}
\label{subsec:midecial_image_translation}

\noindent \textbf{Generative Models for Natural Images.}
Generative models have become fundamental to Image-to-Image (I2I) translation in natural image domains. Conditional GANs such as Pix2Pix~\cite{wang2018highresolutionimagesynthesissemantic, mirza2014conditionalgenerativeadversarialnets} learn direct mappings between paired domains, but are limited by their one-to-one generation strategy. 
Subsequent methods like CycleGAN~\cite{zhu2017unpaired} and DRIT++~\cite{lee2019dritdiverseimagetoimagetranslation} enable diverse outputs via unpaired translation, though GANs still suffer from training instability and mode collapse. 
Diffusion models have emerged as a more stable alternative, offering high-quality synthesis without task-specific tuning, as shown in Palette~\cite{saharia2022palette}, SDEdit~\cite{meng2021sdedit}, and LBM~\cite{chadebec2025lbmlatentbridgematching}. 
Latent-space approaches such as VQ-GAN~\cite{esser2021taming} and LDM~\cite{rombach2022high} improve efficiency and fidelity, while BBDM~\cite{li2023bbdm} further enhances translation stability. 
Despite these advances, most generative methods remain tailored to natural images, with limited adoption in medical imaging contexts.

\noindent \textbf{Medical Image Translation Models.} 
In the medical imaging domain, several studies have proposed GAN-based models for image translation. 
UP-GAN~\cite{upadhyay2021uncertainty} introduces an uncertainty-guided progressive learning strategy to facilitate translation across different imaging modalities. 
RegGAN~\cite{regGAN} incorporates a registration-based adversarial framework that jointly performs image translation and spatial alignment to improve anatomical consistency. 
More recently, diffusion-based models have gained attention in this field. 
SynDiff~\cite{ozbey2023unsupervisedmedicalimagetranslation} employs a conditional diffusion process to generate high-quality medical images, while CPDM~\cite{nguyen2025CPDM} leverages a Brownian Bridge mechanism to directly synthesize PET from CT scans. 
By integrating domain-specific priors into the diffusion process, CPDM enhances the visual quality and clinical utility of the synthesized outputs.


\section{Proposed CardiAC Dataset}

\begin{table}[t]
\centering
\caption{Statistics of the CardiAC dataset (M: Male, F: Female).}
\setlength{\tabcolsep}{3pt}
\resizebox{1.0\linewidth}{!}
{
\begin{tabular}{l|l|l|l|l|l}
\toprule
\textbf{Year} & \textbf{Studies (M, F)} & \textbf{Age} & \textbf{Height (m)} & \textbf{Weight (kg)} & \textbf{\# Slices} \\ 
\midrule
2022 & 186 (139, 47) & 65.72 $\pm$ 10.65 &  1.62  $\pm$ 0.07 & 62.2 $\pm$ 8.99 & 31,788 \\
2023 & 238 (184, 54) & 65.19 $\pm$ 9.86  & 1.62 $\pm$ 0.12 &  63.8 $\pm$ 11.27 & 41,892 \\
\midrule
Total & 424 (323, 101) &  65.42 $\pm$ 10.22 & 1.62 $\pm$ 0.10 & 63.2 $\pm$ 10.51 & 73,680 \\
\bottomrule
\end{tabular}
}
\end{table}

The proposed CardiAC dataset consists of 424 patient studies collected from a large hospital system with multiple branches nationwide.
For each study, six paired NAC and AC cardiac images are provided under both rest and stress conditions, corresponding to three standard orientations: Vertical Long Axis (VLA), Horizontal Long Axis (HLA), and Short Axis (SA).
In addition, each study includes two attenuation maps (rest and stress), derived from low-dose CT scans and used as reference images for attenuation correction during SPECT reconstruction.
All acquisitions are performed on a SPECT system (GE Medical systems, Nuclear) following standard MPI protocols. 
The majority of patients (413 patients) underwent a 2-day imaging protocol, while smaller subsets followed a 1-day stress–rest protocol (9 patients) or a 1-day rest–stress protocol (1 patient). 
One study lacks a specified acquisition description. Each examination includes separate stress and rest acquisitions.
Technetium-99m (Tc-99m) serves as the radiopharmaceutical, with an energy window of either 126–154 keV or 126.45–154.55 keV. 
ECG triggering is applied during acquisition; however, reconstructed DICOM series report Num ECT Phases = 0, indicating that only static (non-gated) perfusion images are retained for analysis. 
Image acquisition employs a low-energy high-resolution parallel-hole collimator in step-and-shoot mode, with one detector head active per reconstruction. 
The acquisition matrix is 128 $\times$ 128, with pixel spacing and slice thickness of approximately 3.2 mm, resulting in 25--49 slices per volume depending on the protocol. 
Attenuation maps are stored as three-dimensional volumes (128 $\times$ 128 $\times$ 128), with each voxel encoding CT-derived linear attenuation coefficients used during reconstruction. 
All image series are reconstructed on GE Xeleris workstations (predominantly version 4.0117, with a minority on earlier versions such as 3.1108), and acquisition console firmware versions include 1.003.429.0 and 1.004.050.15.

\begin{figure*}[t]
    \centering
    \includegraphics[width=\textwidth]{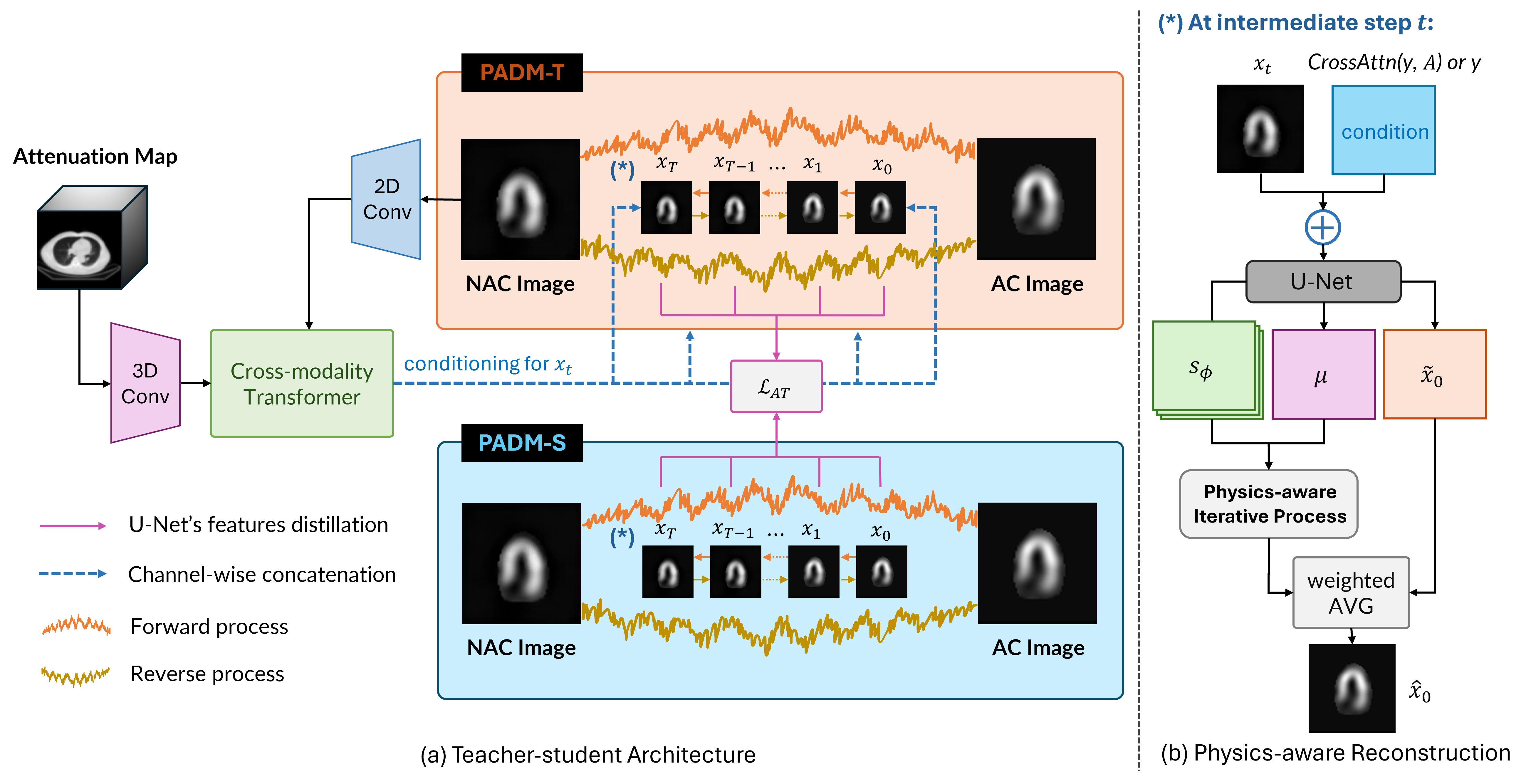} 
    \caption{Overview of the proposed PADM method. 
    (a) Teacher–student framework: The teacher network (PADM-T) conditions the diffusion process on the Attenuation map via a cross-modality transformer, while the student network (PADM-S) learns from NAC images and is guided by feature distillation.
    (b) Physics-aware reconstruction: At each step of the diffusion process, the U-Net predicts projections \(s_\phi\), mean \(\mu\), and clean image \(\tilde{x}_0\), which are refined through a physics-aware iterative update to produce final output.
    }
    \label{fig:proposed_method}
\end{figure*}

\section{Proposed Method}
\label{sec:proposed_method}

\label{subsec:motivation}

\subsection{Motivation}
A common line of research formulates attenuation correction as a direct I2I translation task from NAC-to-AC.  
Formally, given a NAC slice $I_{\text{NAC}} \in \mathbb{R}^{H \times W}$ and its AC counterpart $I_{\text{AC}} \in \mathbb{R}^{H \times W}$, the goal is to learn a mapping as:
\begin{equation}
    f_{\theta}: I_{\text{NAC}} \mapsto I_{\text{AC}}.
\end{equation}
While conceptually straightforward, this formulation is inherently ill-posed: the model must infer the complex, non-linear relationship between tracer distribution and photon attenuation without explicit physical constraints.  
Consequently, direct I2I approaches often suffer from instability, mode collapse, and hallucinated anatomical structures, limiting their clinical applicability and reliability.

\noindent An alternative line of research models attenuation correction indirectly, by first predicting an Attenuation map $\hat{A}$ and then performing physics-based iterative reconstruction~\cite{formula2016}:
\begin{equation}
\hat{A} = g_{\phi}(I_{\text{NAC}}), \quad 
I_{\text{NAC}} \xrightarrow{g_{\phi}} \hat{A} \xrightarrow{\text{Iterative rec.}} I_{\text{AC}}.
\end{equation}
This formula-based paradigm leverages the true acquisition process, providing a principled, physics-informed foundation for reconstruction.  
However, it also highlights three key challenges: (i) the need for stability in generative modeling, as iterative reconstructions are sensitive to errors; (ii) dataset limitations, since not all datasets provide attenuation maps for training; and (iii) alignment issues, with NAC slices and attenuation maps often imperfectly registered on a slice-by-slice basis.

\noindent Motivated by these considerations, we adopt an indirect, physics-guided approach that integrates a stable diffusion backbone and a teacher–student knowledge distillation strategy to address missing attenuation maps and slice misalignment. The goal is to enable accurate and robust attenuation-corrected reconstruction.

\subsection{Overview}
Figure~\ref{fig:proposed_method} presents an overview of the proposed PADM method. 
PADM combines physics-guided priors, knowledge distillation, and generative diffusion modeling, with the generative component inspired by the Brownian bridge diffusion process~\cite{li2023bbdm}. 
The architecture features a teacher--student diffusion framework (PADM-T and PADM-S) for NAC-to-AC synthesis and a physics-aware reconstruction module that leverages CT-derived attenuation maps. 

\vspace{5pt}
\noindent \textbf{Teacher Network.} 
The teacher network $\mathcal{T}_{\theta}$ receives a NAC slice $I_{\text{NAC}} \in \mathbb{R}^{H \times W}$ and the a 3D Attenuation map $A \in \mathbb{R}^{H' \times W' \times D'}$ to predict the corresponding AC slice as:
\begin{equation}
\hat{I}_{\text{AC}}^{\mathcal{T}} \sim \mathcal{T}_{\theta}(I_{\text{NAC}}, A).
\end{equation}
A cross-modality transformer module fuses tracer uptake with attenuation information, compensating for possible misalignment between $I_{\text{NAC}}$ and $A$. The teacher models clinical reconstruction via physics-guided iterative updates, leveraging a Brownian Bridge diffusion backbone to ensure anatomical consistency and reduce generative artifacts.

\vspace{5pt}
\noindent \textbf{Student Network.} 
The student ($\mathcal{S}_\phi$) performs inference using only NAC slices with the same architecture as the teacher. 
It predicts the AC output as:
\begin{equation}
\hat{I}_{\text{AC}}^{\mathcal{S}} \sim \mathcal{S}_{\phi}(I_{\text{NAC}}),
\end{equation}
and is trained via knowledge distillation to replicate the teacher's outputs. 
Architectural consistency between teacher and student facilitates effective transfer of physics-informed representations, enabling the student to produce high-fidelity AC predictions.


\vspace{5pt}
\noindent \textbf{Cross-Modality Transformer Attention.}  
We adopt a Transformer-style cross-attention mechanism to fuse NAC slices with their corresponding attenuation maps.  
For clarity, we define the fusion output as:
\begin{equation}
\small
    X_\text{out} = \text{CrossAttn}(I_{\text{NAC}}, A),
\end{equation}
where $\text{CrossAttn}(\cdot, \cdot)$ denotes the full fusion operation. This process begins by projecting both the NAC image $I_{\text{NAC}}$ and the attenuation map $A$ into latent feature spaces via separate convolutional layers, followed by cross-attention and a feed-forward network:
\begin{equation}
\small
\begin{aligned}
X_\text{NAC} &= \text{Conv}_\text{NAC}(I_{\text{NAC}}), \\
X_A &= \text{Conv}_A(A), \\
\tilde{X}_\text{NAC} 
  &= \text{LayerNorm}\Big(X_\text{NAC} + \text{Attention}(X_\text{NAC}, X_A, X_A)\Big), \\
X_\text{out} 
  &= \text{LayerNorm}\Big(\tilde{X}_\text{NAC} + \text{FFN}(\tilde{X}_\text{NAC})\Big),
\end{aligned}
\end{equation}
with attention defined as $\text{Attention}(Q, K, V) = \text{Softmax}\left(\frac{Q K^\top}{\sqrt{d}}\right) V$.
The resulting features $X_\text{out}$ serve as physics-informed guidance for the teacher model, providing an informative reference to the student model, which is trained without access to attenuation maps.

\vspace{5pt}
\noindent \textbf{2D-to-2D Diffusion Process.}  
We implement at the 2D slice level to ensure pixel-level fidelity, which is critical for clinical applications.  
By combining a Physics-aware Brownian Bridge diffusion process and Teacher-to-Student distillation, we achieve stable and reliable AC reconstruction suitable for practical deployment.

\subsection{Conditional Brownian Bridge Diffusion Process}
\label{subsec:diffusion process}

Inspired by the Brownian Bridge diffusion process~\cite{li2023bbdm}, we adopt it as the diffusion process to map NAC-to-AC slices.
For simplicity, we denote the NAC slice \( I_{\text{NAC}} \) as \( y \in \mathbb{R}^{H \times W} \) and the corresponding AC slice \( I_{\text{AC}} \) as \( x \in \mathbb{R}^{H \times W} \). 
In our implementation, we do not embed images into a VQ-GAN~\cite{esser2021taming} latent space due to the low resolution of the preprocessed NAC and AC slices. 
Instead, the model operates directly in image space, where a network approximates the Physics-aware AC reconstruction at each step of the diffusion process. 
To further improve smoothness and visual fidelity, the U-Net~\cite{ronneberger2015u} is used to predict a refinement image that enhances the final reconstruction.

\noindent \textbf{Forward Process.} 
Following Li et al.~\cite{li2023bbdm}, the forward diffusion maps the AC image $\boldsymbol{x}_0 := x$ to the NAC image $\boldsymbol{y}$. At timestep $t$, the latent state is $\boldsymbol{x}_t = (1-m_t)\boldsymbol{x}_0 + m_t\boldsymbol{y} + \sqrt{\delta_t}\epsilon_t$,  
where $m_t = t/T$, $T$ is the total number of steps, $\delta_t$ is the Brownian Bridge variance, and $\epsilon_t \sim \mathcal{N}(\mathbf{0},\mathbf{I})$.  
The process distribution is as:  
\begin{equation}
\small
    q_{BB}(\boldsymbol{x}_t|\boldsymbol{x}_0,\boldsymbol{y}) 
    = \mathcal{N}\!\big((1-m_t)\boldsymbol{x}_0+m_t\boldsymbol{y}, \delta_t\mathbf{I}\big).
\end{equation}
Throughout the training phase, we employ the following formula to establish the transition probability between two consecutive steps: 
{\small
\begin{align}
    q_{BB} (\boldsymbol{x}_t \mid \boldsymbol{x}_{t-1}, \boldsymbol{y}) &= \mathcal{N} \left( \boldsymbol{x}_t;\,  
    \frac{1 - m_t}{ 1 - m_{t-1} } \boldsymbol{x}_{t-1} \right. \nonumber \\ 
    &\left. + \left(m_t - \frac{1 - m_t}{ 1 - m_{t-1}} m_{t-1}\right) \boldsymbol{y},\,
    \delta_{t \mid t-1} \mathbf{I} \right), \nonumber  \\
    \text{with} \quad 
    \delta_{t \mid t-1} &= \delta_t - \delta_{t-1} \frac{(1 - m_t)^2}{(1 - m_{t-1})^2}.
\label{eq:forward_process}
\end{align}
}

\begin{algorithm}[t]
\caption{Diffusion Training Process}\label{alg:training_proposed_model}
\begin{algorithmic}[1]
\small
\REPEAT
    \STATE Paired data: AC image $\boldsymbol{x_0} \sim q(\boldsymbol{x_0})$, NAC image $\boldsymbol{y} \sim q(\boldsymbol{y})$
    \STATE Timestep \scalebox{0.9}{$t \sim Uniform(1, \dots, T)$}
    \STATE Gaussian noise \scalebox{0.9}{$\epsilon \sim \mathcal{N}(\mathbf{0}, \mathbf{I})$}
    \STATE Forward diffusion \scalebox{0.9}{$\boldsymbol{x_t} =\left(1-m_t\right) \boldsymbol{x_0} +  m_t  \boldsymbol{y}+\sqrt{\delta_t} \epsilon$}
    \STATE Take gradient descent: \\
     \scalebox{0.9}{$\nabla_\theta\left\|m_t\left(\boldsymbol{y} - \boldsymbol{x_0} \right)+\sqrt{\delta_t} \epsilon - (\boldsymbol{x_t} - \boldsymbol{X}_\theta \left(\text{concat}(\boldsymbol{x_t}, \boldsymbol{C}), t\right))\right\|_1$}
     \normalsize
\UNTIL{converged}
\normalsize
\end{algorithmic}
\end{algorithm}

\noindent \textbf{Reverse Process.} 
In the reverse phase, we initialize with $\boldsymbol{x}_T := \boldsymbol{y}$.  
Cross-attention features are computed between the NAC slice $\boldsymbol{y}$ and the Attenuation map ${A}$ as:
\[
\boldsymbol{C} =
\begin{cases}
\text{CrossAttn}(\boldsymbol{y}, {A}) & , \text{teacher model}; \\
\boldsymbol{y} & , \text{student model};
\end{cases}
\]
and concatenated with the latent representation $\boldsymbol{x}_t$ at each diffusion step $t$. The conditional distribution of the reverse transition is then formulated as:
\begin{equation}
\small
\begin{aligned}
    p_\theta(\boldsymbol{x}_{t-1} \mid \boldsymbol{x}_t, \boldsymbol{C}, \boldsymbol{y}) 
    &= \mathcal{N} (\boldsymbol{x}_{t-1}; \mu_\theta (\boldsymbol{x}_t, \boldsymbol{C}, \boldsymbol{y}, t ), \tilde{\delta}_t \mathbf{I}), \\
    \mu_\theta (\boldsymbol{x}_t, \boldsymbol{C}, \boldsymbol{y}, t) 
    &= c_{x t} \boldsymbol{x}_t + c_{y t} \boldsymbol{y}  \\ 
    &\quad + c_{\epsilon t}(\boldsymbol{x}_t - \boldsymbol{X}_\theta (\text{concat}(\boldsymbol{x}_t, \boldsymbol{C}), t)), \nonumber
\label{eq:reverse_process}
\end{aligned}
\end{equation}
where $\mu_\theta(\cdot)$ denotes the estimated mean and $\tilde{\delta}t$ corresponds to the variance of the Gaussian distribution at timestep $t$. The term $\boldsymbol{X}_\theta(\cdot)$ denotes a network that directly predicts the reconstructed AC image $\hat{\boldsymbol{x}}_0$ 
from the noisy latent $\boldsymbol{x}_t$, concatenated with the teacher/student cross-attention features $\boldsymbol{C}$.  
The coefficients $c_{xt}$, $c_{yt}$, and $c_{\epsilon t}$ are fixed quantities, computed directly from $m_t$, $m_{t-1}$, $\delta_t$, and $\delta_{t-1}$ as:
{\small
\begin{align}
    c_{x t} &= \frac{\delta_{t-1}}{\delta_t} \frac{1-m_t}{1-m_{t-1}}+\frac{\delta_{t \mid t-1}}{\delta_t}\left(1-m_{t-1}\right), \nonumber \\
    c_{y t} &= m_{t-1}-m_t \frac{1-m_t}{1-m_{t-1}} \frac{\delta_{t-1}}{\delta_t}, 
    \quad c_{\epsilon t} = \left(1-m_{t-1}\right) \frac{\delta_{t \mid t-1}}{\delta_t}.  \nonumber
\end{align}
}

\noindent \textbf{Diffusion Training Objective.} 
The model is trained to align the predicted distribution with the forward diffusion process (see Algorithm~\ref{alg:training_proposed_model}). Concretely, a neural network parameterized by $\theta$ is used to estimate the mean $\mu_\theta(\boldsymbol{x}_t, \boldsymbol{C}, \boldsymbol{y}, t)$, and the objective is optimized via maximum likelihood by minimizing the Evidence Lower Bound Objective (ELBO).
To obtain a  training objective, we substitute the forward and reverse distributions from Equations~\ref{eq:forward_process} and~\ref{eq:reverse_process}, respectively. 
The loss is computed as:
{\small
\begin{equation}
\begin{aligned}
    \mathcal{L}_\text{ELBO} = \mathbb{E}_{\boldsymbol{x}_0, \boldsymbol{y}, \boldsymbol{\epsilon}}\Big[
    & c_{\epsilon t}\big\| m_t(\boldsymbol{y}-\boldsymbol{x}_0) + \sqrt{\delta_t}\epsilon \\
    & - (\boldsymbol{x}_t - \boldsymbol{X}_\theta(\text{concat}(\boldsymbol{x}_t, \boldsymbol{C}), t))
    \big\|_1 \Big].
\end{aligned}
\end{equation}
}

\noindent \textbf{Sampling Process.} 
We follow the DDIM method~\cite{song2020denoising} for sampling, which accelerates generation by modeling the denoising trajectory as non-Markovian while preserving the same marginal distributions as standard Markovian diffusion (see Algorithm~\ref{alg:sampling_proposed_model}).

\begin{algorithm}[t]
\caption{Sampling Process}\label{alg:sampling_proposed_model}
\begin{algorithmic}[1]
\small
\STATE Sample conditional input: NAC image $\boldsymbol{x_T} = \boldsymbol{y} \sim q(\boldsymbol{y})$;
\FOR{$t=T, \ldots, 1$}
    \STATE \scalebox{0.83}{$\boldsymbol{z} \sim \mathcal{N}(\mathbf{0}, \mathbf{I})$ if $t>1$, else $\boldsymbol{z} = \mathbf{0}$}
    \STATE \scalebox{0.83}{$\boldsymbol{x_{t-1}} =  c_{xt} \boldsymbol{x_t} + c_{yt}\boldsymbol{y} -c_{\epsilon t} (x_t - \boldsymbol{X}_\theta\left(\text{concat}(\boldsymbol{x_t}, \boldsymbol{C}), t\right))+\sqrt{\tilde{\delta}_t} \boldsymbol{z} $}
\ENDFOR
\RETURN $\boldsymbol{x_0}$
\normalsize
\end{algorithmic}
\end{algorithm}

\subsection{Physics-aware Brownian Bridge Diffusion with Learned Path Lengths}
\label{subsec:formula}

At each diffusion step \( t \), the denoising network \( \boldsymbol{X}_\theta \) is redefined to produce an attenuation-corrected reconstruction. 
Specifically, the network output is expressed as the NAC image \( \boldsymbol{y} \) multiplied elementwise by an Attenuation Correction Factor (ACF), parameterized by the network input as:
\begin{equation}
\small
\boldsymbol{X}_\theta\big(\operatorname{concat}(\boldsymbol{x}_t,\boldsymbol{C}),t\big)
= \boldsymbol{y} \odot \mathrm{ACF}_\theta^{(t)}\!\big(\operatorname{concat}(\boldsymbol{x}_t,\boldsymbol{C})\big).
\label{eq:Xtheta_yACF}
\end{equation}
In conventional SPECT reconstruction \cite{formula2016}, the voxel-wise ACF at location \( \mathbf{p}_{i,j,k} \) is defined as:
\begin{equation}
\small
\mathrm{ACF}(i,j,k) =
\left(\frac{1}{N} \sum_{m=1}^N
\exp\!\left[-\,\mu(i,j,k)\, s_{\phi_m}(i,j,k)\right]\right)^{-1},
\label{eq:acf}
\end{equation}
where \( \{\phi_m\}_{m=1}^N \) are the projection angles, \( s_{\phi_m} \) denotes the path length for each angle, and \( \mu \) is the voxel-wise attenuation coefficient.

\noindent In our formulation, the voxel-level quantities \( \{s_{\phi_m}\} \) and \( \mu \) are predicted directly from the network input at diffusion step \( t \), denoted as \( \mathbf{z}^{(t)} = \operatorname{concat}(\boldsymbol{x}_t, \boldsymbol{C}) \). Specifically, a U-Net \( \mathcal{F}_\theta \) maps the concatenated input to a set of per-angle path length fields, an attenuation map, and an auxiliary reconstruction channel as:
\begin{equation}
\small
(\{s_{\phi_m,\theta}^{(t)}\}_{m=1}^N, \mu_\theta^{(t)}, \tilde{x}_0^{(t)})
\;=\;
\mathcal{F}_\theta\!\big(\operatorname{concat}(\boldsymbol{x}_t, \boldsymbol{C})\big).
\label{eq:unet_output}
\end{equation}

\noindent Substituting the network predictions into the voxel-wise definition in Equation~\eqref{eq:acf} yields an input-dependent, parameterized ACF at step \( t \) as:
\begin{equation}
\small
\label{eq:acf_theta}
\begin{split}
\mathrm{ACF}_\theta^{(t)}(\mathbf{p})
&= \Bigg(
     \frac{1}{N}\sum_{m=1}^N
     \exp\big[-\,\mu_\theta^{(t)}(\mathbf{p})\; s_{\phi_m,\theta}^{(t)}(\mathbf{p})\big]
  \Bigg)^{-1}
\end{split}
\end{equation}

\noindent The predicted pair \( \{s_{\phi_m,\theta}^{(t)}\},\,\mu_\theta^{(t)} \) is then passed to a physics-aware
iterative module \( \mathcal{P} \), which explicitly follows the attenuation-correction formulation
in Equation~\eqref{eq:acf_theta} to produce a geometry-consistent reconstruction as:
\begin{equation}
\small
\bar{x}^{(t)} \;=\; \mathcal{P}\Big(\{s_{\phi_m,\theta}^{(t)}\}_{m=1}^N,\; \mu_\theta^{(t)}\Big).
\label{eq:physics_iter}
\end{equation}

\noindent Finally, the refined estimate of the clean image at step \( t \) is obtained by combining the
physics-consistent reconstruction \( \bar{x}^{(t)} \) with the auxiliary channel \( \tilde{x}_0^{(t)} \):
\begin{equation}
\small
\hat{x}_0^{(t)} = \alpha\,\bar{x}^{(t)} + (1-\alpha)\,\tilde{x}_0^{(t)},
\label{eq:refine}
\end{equation}
where \( \alpha \in [0,1] \) is a fixed weighting factor.

\begin{figure*}[t]
    \centering
    \begin{minipage}{0.40\textwidth}
    \vspace{-15pt}
\setlength{\tabcolsep}{3pt}
\captionof{table}{
Comparison of PADM against the baseline diffusion models, i.e., BBDM and BBDM without VQGAN. 
{PADM-T} denotes the teacher model, and {PADM-S} denotes the student model. 
$\downarrow$ indicates lower is better, $\uparrow$ indicates higher is better.  
Diff. (\%) is computed as the relative difference from the baseline score.
\label{tab:metricvariantbbdm}
}
\resizebox{1.0\linewidth}{!}
{
    \begin{tabular}{l||r|r|r||r|r|r}
    \toprule
    \multirow{2}{*}{\textbf{Method}} & \multicolumn{3}{c||}{\textbf{BBDM}} & \multicolumn{3}{c}{\textbf{BBDM w/o VQGAN}} \\
    \cmidrule(lr){2-4}\cmidrule(lr){5-7}
    & \textbf{RMSE} $\downarrow$ & \textbf{SSIM} $\uparrow$ & \textbf{PSNR} $\uparrow$ & \textbf{RMSE} $\downarrow$ & \textbf{SSIM} $\uparrow$ & \textbf{PSNR} $\uparrow$ \\
    \midrule
    Base & 0.0256 & 0.9451 & 33.04 & 0.0330 & 0.9553 & 30.76 \\
    \midrule
    PADM-T & 0.0217 & 0.9796 & 34.98 & 0.0217 & 0.9796 & 34.98 \\
    Diff (\%) & \colorbox{green!30}{+15.2\%} & \colorbox{green!30}{+3.7\%} & \colorbox{green!30}{+5.9\%} & \colorbox{green!30}{+34.2\%} & \colorbox{green!30}{+2.6\%} & \colorbox{green!30}{+13.7\%} \\
    \midrule
    PADM-S & 0.0218 & 0.9795 & 34.75 & 0.0218 & 0.9795 & 34.75 \\
    Diff (\%) & \colorbox{green!30}{+14.8\%} & \colorbox{green!30}{{+3.7\%}} & \colorbox{green!30}{+5.2\%} & \colorbox{green!30}{+33.9\%} & \colorbox{green!30}{+2.6\%} & \colorbox{green!30}{+13.0\%} \\
    \bottomrule
    \end{tabular}
}
\vspace{0.5cm}
    \end{minipage} \hfill
    \begin{minipage}{0.27\textwidth}
    \vspace{-7.5pt}
        \centering
    \centering
    \small
    \setlength{\tabcolsep}{3pt}
    \captionof{table}{Comparison of PADM across different numbers of projections. 
    Proj. is projections.
    Diff. (\%) denotes the relative performance gap of the student (PADM-S) compared to the teacher (PADM-T).
    ~\label{tab:teacher_student}}
    \resizebox{1.0\linewidth}{!}
    {
\begin{tabular}{c|l|r|r|r}
\toprule
\textbf{Proj.} & \textbf{Method} & \textbf{RMSE} $\downarrow$ & \textbf{SSIM} $\uparrow$ & \textbf{PSNR} $\uparrow$ \\
\midrule
\multirow{3}{*}{16}
    & PADM-T & 0.0217 & 0.9796 & 34.98 \\
    & PADM-S & 0.0218 & 0.9795 & 34.75 \\
    & {Diff. (\%)} & \colorbox{red!30}{-0.46\%} & \colorbox{red!30}{-0.01\%} & \colorbox{red!30}{-0.65\%} \\
    \midrule
    \multirow{3}{*}{32}
    & PADM-T & 0.0217 & 0.9796 & 34.98 \\
    & PADM-S & 0.0221 & 0.9806 & 34.59 \\
    & {Diff. (\%)} & \colorbox{red!30}{-1.84\%} & \colorbox{red!30}{-0.10\%} & \colorbox{red!30}{-1.11\%} \\
    \midrule
    \multirow{3}{*}{64}
    & PADM-T & 0.0218 & 0.9796 & 34.99 \\
    & PADM-S & 0.0229 & 0.9777 & 34.16 \\
    & {Diff. (\%)} & \colorbox{red!30}{-5.04\%} & \colorbox{red!30}{-0.19\%} & \colorbox{red!30}{-2.37\%} \\
\bottomrule
\end{tabular}

    }
    \end{minipage} \hfill
    \begin{minipage}{0.295\textwidth}
    \vspace{-10pt}
    \centering
    \small
    \captionof{table}{
    Comparison of PADM with others methods on our CardiAC dataset.  
    The \textcolor{red}{best} and \textcolor{blue}{second best} results are highlighted in the red and blue. 
    {Diff. (\%)} shows the relative performance gaps of PADM compared to the nearest methods. 
    \label{tab:cpdm_results}
    \vspace{-10pt}
    }
    \setlength{\tabcolsep}{3pt}
    \resizebox{1.0\linewidth}{!}
    {
        \begin{tabular}{l|r|r|rr}
        \toprule
        \textbf{Method} & \textbf{RMSE $\downarrow$} & \textbf{SSIM $\uparrow$} & \textbf{PSNR $\uparrow$} \\
        \midrule
        Pix2Pix \cite{isola2017image}  & 0.0269 & 0.9601 &  32.59 \\
        RegGAN \cite{regGAN} & \textcolor{blue}{0.0253} & 0.9797 & \textcolor{blue}{33.46} \\
        ResViT \cite{Dalmaz_2022}  & 0.0266 & \textcolor{blue}{0.9840} & 32.91 \\
        UNIT \cite{liu2017unsupervised} &  0.0345 & \textcolor{red}{0.9848} & 30.28 \\
        BBDM \cite{li2023bbdm}    & 0.0256  & 0.9451 & 33.04 \\
        Palette \cite{saharia2022palette} & 0.0608 & 0.6455 & 26.64 \\
        \midrule
        PADM  & \textcolor{red}{0.0218} & 0.9795 & \textcolor{red}{34.75} \\
        \midrule
        {Diff. (\%)} & \colorbox{green!30}{+13.83\%} & \colorbox{red!30}{-0.53\%} & \colorbox{green!30}{+3.82\%}\\
        \bottomrule
        \end{tabular}
    }
    \end{minipage}
    \vspace{-0.2cm}
\end{figure*}

\subsection{Teacher-to-Student Knowledge Distillation} 
\label{subsec:kd}
{
To transfer knowledge from the teacher network, which is conditioned on Attenuation maps, to the student network, we employ Attention Transfer (AT). 
Let the teacher and student compute aggregated attention maps as:
\begin{equation}
\small
{Q}_T = \text{Agg}(\boldsymbol{C}_\text{T}), \quad 
{Q}_S = \text{Agg}(\boldsymbol{C}_\text{S}),
\end{equation}
where $\boldsymbol{C}_\text{T} = \text{CrossAttn}(\boldsymbol{y}, {A})$ denotes the teacher's cross-attention features conditioned on the Attenuation map \( A \), and $\boldsymbol{C}_\text{S} = \boldsymbol{y}$ denotes the student's attention features without conditioning.
The student aligns its normalized attention with the teacher as:
\begin{equation}
\small
\mathcal{L}_\text{AT} = 
\left\| \frac{{Q}_\text{T}}{\|{Q}_\text{T}\|_2} - \frac{{Q}_\text{S}}{\|{Q}_\text{S}\|_2} \right\|_2.
\end{equation}
During training, paired NAC and AC slices are fed to both networks. 
The teacher remains frozen, while the student parameters are optimized. The overall objective combines the diffusion loss with the attention transfer loss as:
\begin{equation}
\small
\mathcal{L}_\text{Total} = \mathcal{L}_\text{ELBO} + \lambda \mathcal{L}_\text{AT},
\end{equation}
where \( \lambda \) is a balancing hyperparameter.
}

\section{Experimental Results}

{
In this section, we evaluate the performance of the proposed PADM method using our CardiAC dataset. 
We compare PADM against general-purpose and medical-specific image translation methods, including GAN-based models, i.e., Pix2Pix~\cite{isola2017image}, ResViT~\cite{Dalmaz_2022}, Reg-GAN~\cite{regGAN}, and UNIT~\cite{liu2017unsupervised}; and diffusion-based models, i.e., Palette~\cite{saharia2022palette} and BBDM~\cite{li2023bbdm}.
}

\subsection{Experimental Settings}
\noindent \textbf{Data Preparation.}
{
The dataset is divided into training, validation, and test subsets comprising 254, 84, and 86 patients, respectively. 
All images are normalized to the range $[-1, 1]$. 
Each NAC and AC slice is cropped to a \(50 \times 50\) region of interest, background regions are standardized, and the resulting images are resized to \(256 \times 256 \times 1\). 
Attenuation maps are similarly normalized to $[-1, 1]$ to maintain consistency with the reconstructed SPECT images.
}

\vspace{5pt}
\noindent \textbf{Models \& Hyperparameters.} 
{
We implement the proposed PADM method as described in Section~\ref{sec:proposed_method}. 
The diffusion process follows a Brownian bridge formulation with 500 timesteps during training. 
We use the Adam optimizer~\cite{kingma2014adam} with an initial learning rate of $1\text{e}{-4}$, scheduled via step decay, and a batch size of 8. 
All experiments are conducted on a single NVIDIA RTX A6000 GPU with 48\,GB of VRAM.
}
\begin{figure*}[h]
\centering
\setlength{\tabcolsep}{0pt}
\includegraphics[width=0.95\textwidth]{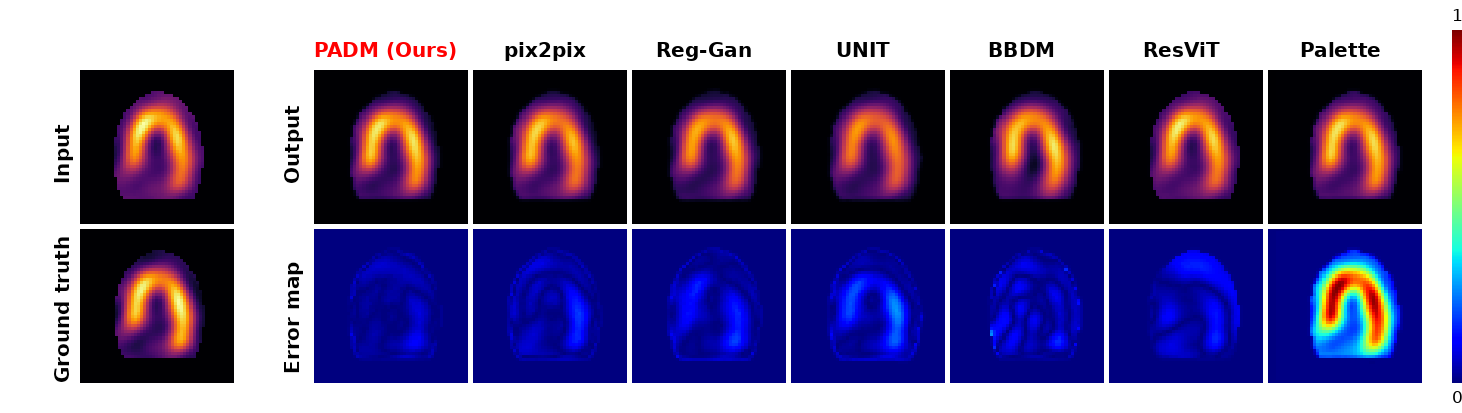} 
\includegraphics[width=0.95\textwidth]{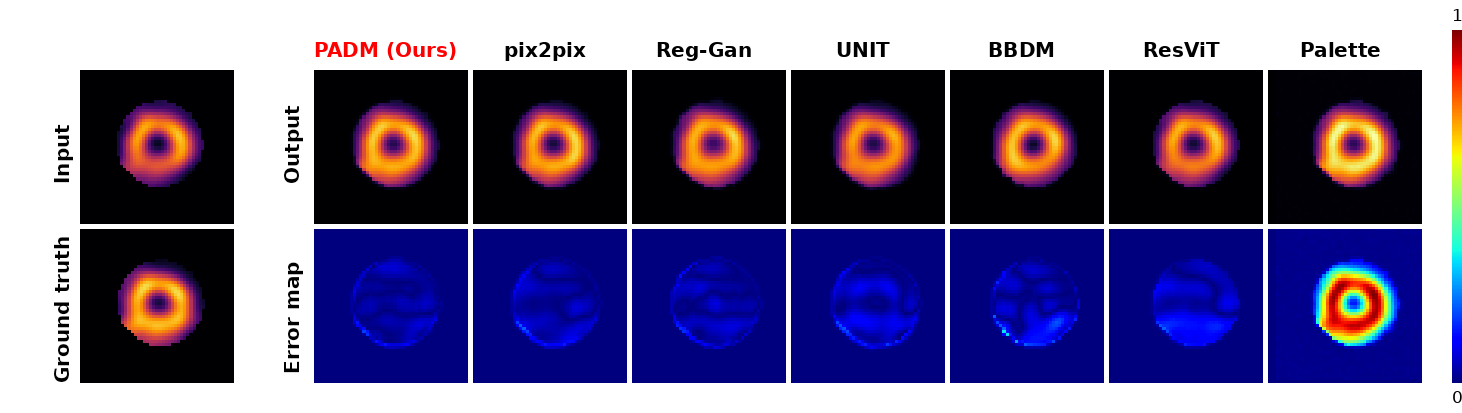}  
\includegraphics[width=0.95\textwidth]{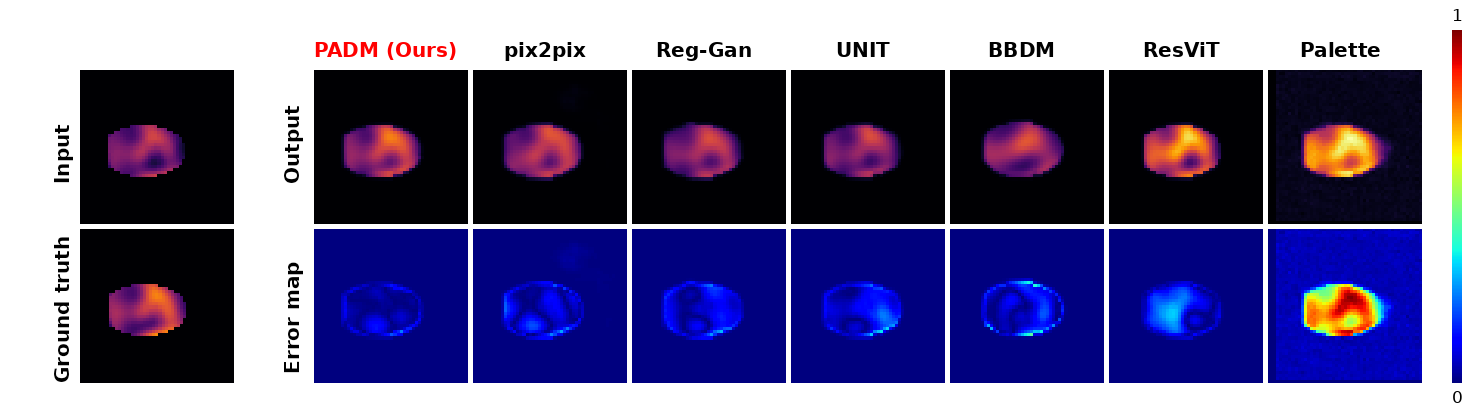}
\caption{Qualitative comparison of reconstructed images across three standard views: horizontal long axis (top), short axis (middle), and vertical long axis (bottom).}
\label{fig:demo_methods}
\vspace{-0.3cm}
\end{figure*}

\vspace{5pt}
\noindent \textbf{Evaluation Metrics.} 
{
We evaluate the quality of generated AC images using Root Mean Square Error (RMSE), Structural Similarity Index Measure (SSIM), and Peak Signal-to-Noise Ratio (PSNR). 
RMSE captures the average magnitude of pixel-wise errors, indicating overall reconstruction accuracy. 
SSIM assesses perceptual quality by measuring structural similarity between the generated and ground-truth images. 
PSNR quantifies pixel-level fidelity, with higher values indicating better visual quality.}

\subsection{Preliminary Analysis}
{
We begin by evaluating the effectiveness of the proposed PADM method under two settings: (1) comparison with baseline diffusion models (Table~\ref{tab:metricvariantbbdm}) and (2) analysis of student--teacher performance across varying numbers of projections used to transfer knowledge from teacher to student (Table~\ref{tab:teacher_student}).
}

\vspace{5pt}
\noindent \textbf{PADM outperforms baseline diffusion models.} 
{
In Table~\ref{tab:metricvariantbbdm}, PADM-T and \mbox{PADM-S} consistently outperform BBDM and its ablated variant without VQGAN across all evaluation metrics. 
Although VQGAN is commonly used as a perceptual compressor to enable diffusion models to operate in a lower-dimensional latent space, the proposed PADM operates directly in image space without VQGAN and still achieves the best performance. 
PADM-T achieves a 15.2\% and 34.2\%  improvement in RMSE compared to BBDM and BBDM without VQGAN, respectively, while \mbox{PADM-S} maintains comparable performance with only marginal degradation. 
}

\vspace{5pt}
\noindent \textbf{PADM is robust to fewer projections.}
{
Table~\ref{tab:teacher_student} shows the performance of PADM across varying numbers of projections transferred from teacher to student. As the number of projections increases from 16 to 64, the student model (PADM-S) exhibits a gradual decline in performance. The RMSE gap widens from \(-0.46\%\) to \(-5.04\%\), and PSNR drops by up to \(-2.37\%\), indicating more pronounced reconstruction errors at higher projection counts. While SSIM varies only slightly, the overall trend suggests that PADM-S struggles to match PADM-T as projection complexity increases. This highlights a potential trade-off in student generalization when scaling up the number of views.}

\subsection{Comparison with Existing Methods}

\noindent \textbf{Quantitative Results.} 
{
Table~\ref{tab:cpdm_results} presents a results of PADM with existing methods on the proposed CardiAC dataset. 
PADM achieves the lowest RMSE (0.0218) and highest PSNR (34.75), outperforming the next-best methods by 13.83\% and 3.82\%, respectively. 
While its SSIM (0.9795) is slightly below the highest score from UNIT (0.9848), PADM remains competitive across all metrics. 
Unlike previous diffusion methods that rely on VQGAN-based latent spaces, PADM operates directly in the image domain and leverages a physics-aware reconstruction strategy. 
By explicitly incorporating imaging geometry into the diffusion process, PADM improves pixel-level accuracy and perceptual quality, particularly in clinically critical regions.
}

\vspace{5pt}
\noindent \textbf{Qualitative Results.} 
{
Figure~\ref{fig:demo_methods} visually compares reconstructed attenuation-corrected images produced by PADM and six baseline methods across three standard cardiac views. 
PADM outputs exhibit higher visual fidelity and sharper anatomical structures, closely matching the ground truth. 
The error maps highlight that PADM consistently produces fewer artifacts and lower residual errors, especially in clinically important regions such as the myocardium. 
In contrast, alternative methods (i.e., Palette, UNIT) show noticeable distortions or elevated error responses. 
These results further validate PADM's superior reconstruction quality and its robustness across different anatomical perspectives.
}

\section{Conclusion}

In this work, we addressed the challenge of attenuation artifacts in cardiac SPECT myocardial perfusion imaging by introducing a new dataset and a novel reconstruction method. 
Specifically, we introduced CardiAC, a dataset that provides paired NAC and AC reconstructions alongside CT-derived attenuation maps, offering a valuable benchmark for future research.
Additionally, we proposed PADM, which integrates explicit physical priors through a teacher--student distillation framework. 
PADM enables accurate NAC-to-AC reconstruction without requiring CT-based Attenuation map input during inference. 
Extensive quantitative and qualitative evaluations show that PADM consistently outperforms existing generative methods in both reconstruction accuracy and perceptual fidelity.


\noindent In future work, we plan to collaborate with clinicians to rigorously evaluate the diagnostic quality of reconstructed images and explore their potential in downstream clinical applications such as lesion classification and report generation, following recent advances in multimodal vision--language modeling for medical imaging~\cite{nguyen2025VLF}.




\section*{Acknowledgment}
This research is funded by Hanoi University of Science and Technology (HUST) under grant number T2024-TĐ-002.

{
    \small
    \bibliographystyle{ieeenat_fullname}
    \bibliography{main}
}

\end{document}